# Lexical Based Semantic Orientation of Online Customer Reviews and Blogs

Aurangzeb khan[1], Khairullah khan[1], Shakeel Ahmad[2], Fazal Masood Kundi[2], Irum Tareen[2], Muhammad Zubair Asghar[2]

[1]Institute of Engineering and Computing Sciences, University Of Science and Technology Bannu, Pakistan.
[2]Institute of Computing and Information Technology, Gomal University D.I.khan, Pakistan.
zubair@gu.edu.pk

___________________________________________________________________________________________

**Abstract**

Rapid increase in internet users along with growing power of online review sites and social media has given birth to sentiment analysis or opinion mining, which aims at determining what other people think and comment. Sentiments or Opinions contain public generated content about products, services, policies and politics. People are usually interested to seek positive and negative opinions containing likes and dislikes, shared by users for features of particular product or service. This paper proposed sentence-level lexical based domain independent sentiment classification method for different types of data such as reviews and blogs. The proposed method is based on general lexicons i.e. WordNet, SentiWordNet and user defined lexical dictionaries for semantic orientation. The relations and glosses of these dictionaries provide solution to the domain portability problem. The method performs better than word and text level corpus based machine learning methods for semantic orientation. The results show the proposed method performs better as it showsprecision of 87% and83% at document and sentence levels respectively for online comments.

**Keywords**: sentiment analysis, opinion mining, classification, semantic orientation,

## Introduction

Sentiment analysis or opinion mining is a sub-discipline within text mining, to identify subjectivity, sentiments, affects and other states of emotions within the text found in the other online resources. Opinion mining is in reference to computational techniques utilized to extract, assess, understand and classify the numerous opinions that are expressed in a variety of online social media comments, news sources and other content created by the user (Chen and Zimbra, 2010). Sentiment is a view, feeling, opinion or assessment of a person for some product, event or service. Sentiment Analysis or Opinion Mining is a challenging Text Mining and Natural Language Processing problem for automatic extraction, classification and summarization of sentiments and emotions expressed in online text (Pang, Lee and Vaithyanathan, 2002)(Hu and Liu, 2004)(Asghar, Khan, Ahmad and Kundi, 2013). Sentiment analysis is replacing traditional and web based surveys conducted by companies for finding public opinion about entities like products and services. Sentiment Analysis also assists individuals and organizations interested in knowing what other people comment about a particular product, service topic, issue and event to find an optimal choice for which they are looking for. By the end of 2011, over 181 million blogs were tracked with 6.5 million personal blogs and 12 million blogs written on social networks with majority of users seeking opinions on products and services(Balahur, Steinberger, Kabadjov, Zavarella, Van Der Goot, Halkia and Belyaeva, 2013)(Andreevskaia, and Bergler, 2008).Sentiment analysis is of great value for business intelligence applications, where business analysts can analyze public sentiments about products, services, and policies (Funk, A., Li, Saggion, Bontcheva, and Leibold, 2008). Sentiment Analysis in the context of Government Intelligence aims at extracting public views on government policies and decisions to infer possible public reaction on implementation of certain policies(Stylios, Christodoulakis, Besharat, Vonitsanou, Kotrotsos, Koumpouri and Stamou, 2010). Feature based sentiment analysis include feature extraction, sentiment prediction, sentiment classification and optional summarization modules (Hu, M., and Liu, 2004). Feature extraction identifies those product aspects which are being commented by customers, sentiment prediction identifies the text containing sentiment or opinion by deciding sentiment polarity as positive, negative or neutral and finally summarization module aggregates the results obtained from previous two steps. Feature extraction process takes text as input and generates the extracted features in any of the forms like Lexico-Syntactic or Stylistic, Syntactic and Discourse based (Ohana, B., 2009).Sentiment analysis allows for a better understanding of customers' feelings regarding various companies, their products and services or the way they handle customer services, as well as the behavior of their individual agents. It can be used to help in customer relationship management, employees training, identifying and resolving difficult problems as they appear (Asghar, Khan, Kundi, Qasim, Khan, RahmanUllah, Nawaz, 2014)(Pang, and Lee, 2008). This work, presents an online customer reviews classification method extracting sentiments from blogs comments. The results describes, that the method performs better, as compared to other techniques.

**Background And Related Work**

Sentiment analysis deals with online customer reviews, blog comments, and other related online social network contents (views and news). People recognizing the usefulness, in the immense expansion of the Web, are being drawn more and more towards online services like, shopping, e-banking, e-commerce etc. as well as to the feedback given in the form of reviews and comments about various products and services. Online reviews and comments added on a daily basis to various online sites, like epinion.com, cnet.com, amazon.com, facebook.com, and twitter.com are quite helpful for consumers in making decisions and for companies planning market strategies (Zhu and Zhang, 2010). This has attracted a lot of attention of research communities from industries as well as academia. Consequently, the steady flow of interest towards online resources in recent times has resulted in a tremendous amount of research activity in the field of sentiment analysis and opinion mining(Liu, 2010a). This has led to the appearance of Web 2.0 which, combined with the vast social media content, has caused quite a bit of excitement as it provide ample opportunities to get a better understanding of what the general public, especially consumers, think about company strategies, product preferences and marketing campaigns as well as social events and political movements. Analysis of the thousands, possibly millions, of reviews, comments and other feedbacks expressed in various forums (Yahoo Forums), blogs (blogosphere), social network and social media sites (including You Tube, Flikr, and Facebook, Twitter etc) and virtual worlds (like Second Life) can potentially answer the numerous new and interesting research questions regarding social, economical, cultural, geo political and business issues (Chen, and Zimbra, 2010).

The early work of sentiment analysis began with subjectivity detection, dating back to the late 1990's. Afterward, it shifted its focus towards the interpretation of metaphors, point of views, narrations, affects, evidentiality in text and other related areas. With the increase in internet usage, the Web became an important source of text repositories. Consequently, a switch was slowly made away from the use of subjectivity analysis and towards the use of sentiment analysis of the Web content. The early work in subjective and objective analysis and classification, the separating of subjective, objective and natural sentences, was a very hard task (Pang, and Lee, 2008)(Turney, 2002)(Wiebe, 1994)(Wiebe, 1990).

In sentence level sentiment analysis, the text document or reviews are split into sentences and each sentence is checked for its semantic orientation by using lexical or statistical techniques. It can be associated with two tasks. The first of these two tasks is to identify whether the sentence is subjective or objective. In the second task, subjective sentences are classified into positive, negative or neutral polarity. Sentence level semantic orientation is important because it takes each sentence individually for semantic orientation. NLP methods are useful for such types of semantic orientations. Sentence level analysis decides what the primary or comprehensive semantic orientation of a sentence is while the primary or comprehensive semantic orientation of the entire document is handled by the document level analysis (Pang, and Lee, 2008)(Hu, and Liu, 2004).In addition to sufficient work being performed in text analytics, feature extraction in sentiment analysis is now becoming an active area of research. A review paper presented by (Asghar, Khan, Ahmad and Kundi, 2014) discusses existing techniques and approaches for feature extraction in sentiment analysis and opinion mining. In this review, the main focus is on state-of-art paradigms used for feature extraction in sentiment analysis. Further evaluation of existing techniques is done and challenges to be solved in this area are addressed.

Many approaches have been adopted for performing sentiment analysis on social media sites. Knowledge based approaches classify the sentiments through dictionaries defining the sentiment polarity of words and linguistic patterns (Asghar, Qasim, Ahmad, Ahmad, Khan, Khan, 2013).However, the text documents or reviews are broken down into sentences for sentiment analysis at the sentence level. These sentences are then evaluated by utilizing lexical or statistical methods in order to determine their semantic orientation. This process involves two functions; first is to determine the subjectivity or objectivity of a sentence and the next function is of taking the subjective sentences for an opinion orientation. Some existing work involves analysis at different levels. Particularly, the level of semantic orientation involving words regarding opinion as well as the phrase level. Semantic orientation can be accumulated from the words and phrases to find out the overall Semantic Orientation of a particular sentence or review(Leung, and Chan, 2008.)(Westerski, 2007)(Hu, and Liu, 2004)(Kundi, Ahmad, Khan and Asghar, 2014)(Andreevskaia, and Bergler, 2008)(Liu, 2010b). A rule based subjectivity classifier, capable of mining user tweets shared on twitter during some key political event, was designed to isolate subjective and objective sentences (Asghar, RahmanUllah, Ahmad, Khan, Ahmad and Nawaz, 2014). The framework for subjectivity and objectivity classification is compatible with both annotated and un-annotated dataset.

However sentence level semantic orientation along with extraction of sentence sense using WSD is not being considered so far. This work, presents a technique for sentence level rule based method for sentiments classification considering semantics strength of sentence from blogs and online reviews. The strength of each sentence in a review is obtained considering all parts of speech.

**Materials and Methods**

In this section, rule based sentiment classification method of online reviews and blogs comments arepresented. The following four steps describe the overall process for semantic orientation for different genre and domains using sentence level lexical dictionaries.
1) Collecting data (text), processing and removal of noise form text data.
2) Developing and using knowledge base which is the collection of lexical dictionaries.
3) Processing of text data at sentence level using WSD for extraction of sentence sense.
4) Checking the polarity of each sentence according to sentence structure and deciding about its opinion orientation (positive, negative or neutral).

This work creates a combination of dictionaries called knowledge base which conations SentiWordNet, WordNet and predefined intensifier dictionaries for rule based polarity classification of positive, negative and neutral opinions. It combines and interlinks the lexical dictionaries (WordNet, SentiWordNet, intensifiers etc.) to make a knowledge base and extract the sense of terms, and semantic score, as described in the next section (Esuli and Sebastiani, 2006). The dictionary database information is described as follows**.**

* The pair (POS, offset) uniquely identifies a WordNet synset. Numeric ID called offset associated with POS uniquely identified a synset in a database.
* The values PosScore and NegScore are the positive and negative scores assigned by SentiWordNet to the synset
* The objectivity score can be calculated as :

$$pos(s) + Neg(s) + Obj(s) = 1 \qquad (1)$$

$$ObjScore = 1 - (PosScore + NegScore) \qquad (2)$$

* Last column of the dictionary database includes synsets (separated by spaces) and gloss information associated with the term.

(Where NegScore= negative Score, PosScore=Positive Score, Pos(s) = Positive score of synset s., Neg(s) = Negative score of synset s., Obj(s) = Objectiveness score of synset s.)

**Results and Discussions**

For evaluation of the proposed method, public feedbacks in term of comments are extracted from cricinfo[1]; the 2011 blog for cricket worldcup. **Table-1** shows the blog comment dataset information. The reviews are spilt into sentences and after separation of sentences, thelexical terms for semantic orientation were extracted.Furthermore, the Part of Speech (POS) tagger was applied to classify the sentences into subjective and objective sentences. The subjective sentences are considered for further processing to find the semantic orientation at the individual sentence level.

Table1: Sum of Opinion Sentences

| Datasets | Comment | Sentence | Subjective | Objective | Percent |
|---|---|---|---|---|---|
| Cricket World Cup 2011 | 500 | 1630 | 1238 | 392 | 76/24 |
| Airlines Reviews | 1000 | 7730 | 5405 | 2325 | 70/30 |

---

[1] http://www.cricinfo.com

To evaluate the proposed method157 Cricket blog feedbacks are taken from www.cricinfo.com as a dataset. The dataset is split into 592 sentences which are manually evaluated for positive, negative and neutral sentiments. Out of these manuallyevaluated sentences, 266 are labelled as positive, 206 as negative and 120 as neutral sentences as shown in **Table-2.** When the proposed method is evaluated on this dataset for sentiment orientation, an accuracy of 83% is achieved at sentence level.

The blog comments of the above dataset are manually evaluated for performance checking at the feedback level. Among

Table 2: Sentiment Orientation of Cricket Blog Comments at Sentence Level

| | | Actual Orientation | | | |
|---|---|---|---|---|---|
| | | Positive | Negative | Neutral | Total |
| System Assigned | Positive | 222 | 26 | 12 | 260 |
| | Negative | 30 | 170 | 8 | 208 |
| | Neutral | 14 | 10 | 100 | 124 |
| | Total | 266 | 206 | 120 | 592 |
| Overall Accuracy | | 0.83 | | | |

157 feedbacks, 86 comments as whole feedback are judged as positive, 53 as negative and 18 as neutral feedbacks as described in

**Table-3.**The objective of this work is to evaluate the capability of the proposed method to correctly classify the semantic orientationof sentences and also to access the positive, negative or neutral sentiments from the dataset. The proposed method achieved 87% results at the feedback level from the sports blog. It is observed that number of sentences in blogs can affect the accuracy at feedback level. If feedback or blog contains more sentences, its accuracy could be higher compared to those having less number of sentences.

Table 3: Sentiment Orientation of Cricket Blog Comments at Feedback Level

| | | Actual Orientation | | | |
|---|---|---|---|---|---|
| | | Positive | Negative | Neutral | Total |
| System Assigned | Positive | 80 | 7 | 3 | 90 |
| | Negative | 5 | 44 | 2 | 51 |
| | Neutral | 1 | 2 | 13 | 16 |
| | Total | 86 | 53 | 18 | 157 |
| Overall Accuracy : | 0.87 | | | | |

The results of proposed method were compared with other machine learning methods presented by (Go, Bhayani and Huang, 2009) and (Andreevskaia, Bergler, 2008)who achieved the accuracy of 80% for classifying positive and negative sentiments. The proposed method describes that pre-processing is more important to remove noisy text in the case of short messages and comments to achieve high accuracy. (Shamma, D. A., Kennedy and Churchill, 2009, October)investigated the twitter blogs comments for the 2008 American Presidential Electoral debates. They illustrated that the analysis of twitter usage is important and closely yield the semantic structure and contents of the media objects. The twitter can be a predictor of the change in any media event. So mining blogs comments play an important role that can be leveraged to evaluate and analyse any activity.The method is compared withthe methods proposed by (Andreevskaia, and Bergler, 2008, February)(Go, Bhayani and Huang, 2009); the proposed method achieved better results than this approach as shown in **Table-4 and Fig-1.**

Table 4: Compression with Other Related Works on Blog Datasets

|  |  | Andreevskaia Bergler,(2008) | Go, Bhayani, Huang,(2009) | Proposed Method |
|---|---|---|---|---|
| **Sentiment Orientation at** | Sentence | 71 | 80 | 83 |
|  | Feedback | 82 | 82 | 87 |

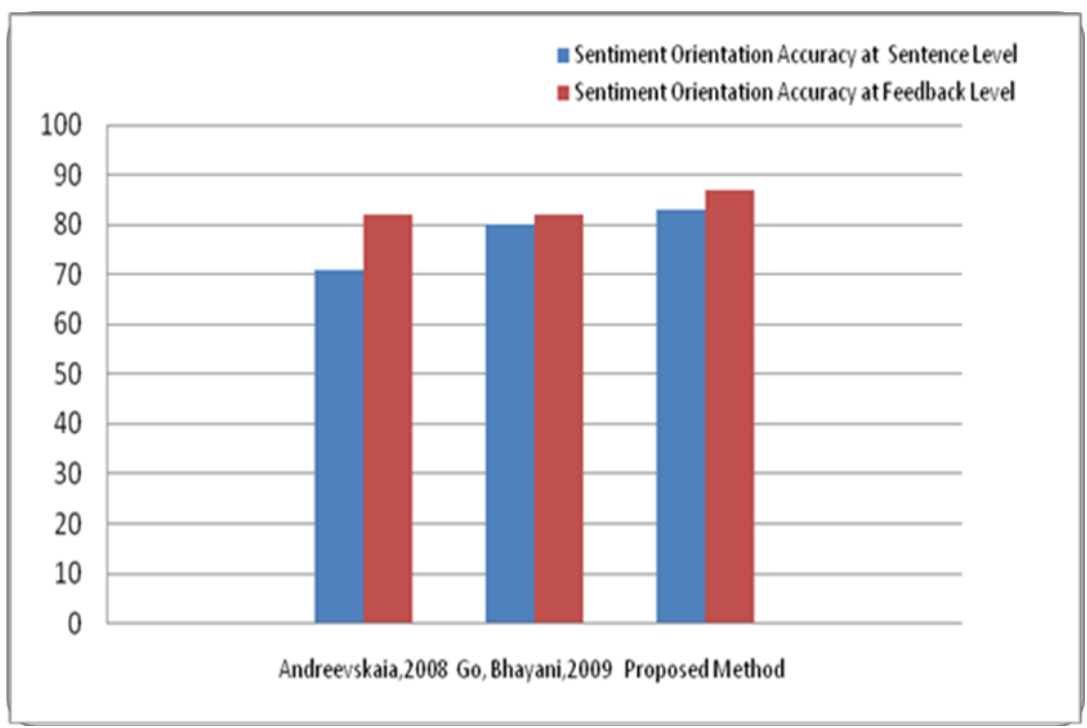

Figure 1: Comparison of proposed method with other techniques for blogs comments

**Conclusion and Future Work**

In this work, a sentence level rule based classification method of online reviews and blogs comments is introduced.
In this paper the process of developing and using knowledge base which is the collection of lexical dictionaries is presented for processing of text data at sentence level using WSD for extraction of sentence sense. The polarity of each sentence is checked according to sentence structure and deciding about opinion orientation (positive, negative or neutral).
From the results the proposed method performs better as compared to other methods as it is clear that the proposed method achieves an accuracy of 83% and 87% at the sentence and document level respectively for blogs short comments.
In future, extraction of the acute sense of sentence and remove noisy text for an efficient semantic orientation. Furthermore, the knowledgebase need to improve for the semantic scores of all parts of speech.